\documentclass[10pt,twocolumn,letterpaper]{article}

\usepackage{multirow}
\usepackage{caption}
\usepackage[accsupp]{axessibility}  

\usepackage{wacv}              

\definecolor{wacvblue}{rgb}{0.21,0.49,0.74}
\usepackage[pagebackref,breaklinks,colorlinks,allcolors=wacvblue]{hyperref}

\def\wacvPaperID{6} 
\def\confName{WACV}
\def\confYear{2026}

\title{S3-CLIP: Video Super Resolution for Person-ReID}

\author{Tamás Endrei\\
Pázmány Péter Catholic University, Hungary\\
{\tt\small endrei.tamas@itk.ppke.hu}
\and
György Cserey\\
Pázmány Péter Catholic University, Hungary\\
{\tt\small cserey.gyorgy@itk.ppke.hu}
}

\begin{document}
\maketitle
\begin{abstract}


Tracklet quality is often treated as an afterthought in most person re-identification (ReID) methods, with the majority of research presenting architectural modifications to foundational models. Such approaches neglect an important limitation, posing challenges when deploying ReID systems in real-world, difficult scenarios. In this paper, we introduce S3-CLIP\footnote{\url{https://github.com/TomasDelaney/S3-CLIP}}, a video super-resolution-based CLIP-ReID framework developed for the VReID-XFD challenge at WACV 2026. The proposed method integrates recent advances in super-resolution networks with task-driven super-resolution pipelines, adapting them to the video-based person re-identification setting. To the best of our knowledge, this work represents the first systematic investigation of video super-resolution as a means of enhancing tracklet quality for person ReID, particularly under challenging cross-view conditions. Experimental results demonstrate performance competitive with the baseline, achieving $37.52\%$ mAP in aerial-to-ground and $29.16\%$ mAP in ground-to-aerial scenarios. In the ground-to-aerial setting, S3-CLIP achieves substantial gains in ranking accuracy, improving Rank-1, Rank-5, and Rank-10 performance by $11.24\%$, $13.48\%$, and $17.98\%$, respectively.

\end{abstract}
    
\section{Introduction}
\label{sec:intro}

Person Re-Identification (ReID) is the task of recognizing and matching pedestrians across non-overlapping camera views, based on images or video tracklets. Recently, this setting has been pushed to the extreme by emerging datasets that introduce severe viewpoint, scale, and distance discrepancies across ground-based and aerial camera systems \cite{zhang2023ground, zhang2024cross, nguyen2025ag, hambarde2025detreidx}. Most modern ReID methods rely on foundational models and their variants such as CLIP \cite{radford2021learning} or DINO \cite{caron2021emerging, oquab2023dinov2}. These state-of-the-art (SOTA) methods \cite{chen2023beyond, yu2024tf, khalid2025bridging} rely on extrapolation techniques such as bicubic or bilinear interpolation, focusing on architectural advances with little consideration for improving input data quality.

Traditional ReID methods focus on challenges such as pose changes, occlusions, or lighting, but routinely assume comparable image quality and scale \cite{jiao2018deep}. In practice, however, surveillance footage often yields very low-resolution (LR) person crops (e.g. due to distance or zoom), while enrollment galleries contain higher-resolution (HR) images \cite{jiao2018deep}. This resolution mismatch severely degrades matching accuracy for example baseline ReID model can lose about 19.2\% rank-1 accuracy when evaluated across such cross-resolution conditions \cite{li2019recover, cheng2020inter}. Motivated by this, there have been attempts to integrate image super-resolution (SR), into the ReID pipeline \cite{li2019recover, adil2020multi, zhang2021deep, han2021adaptive}. Super-resolution can, in principle, recover fine-grained details (clothing texture, facial cues) missing in LR data \cite{jiao2018deep} and align the feature distributions of queries and gallery images. In addition, it can help improve the performance of downstream tasks, which is preserved in out-of-distribution domains \cite{sundaram2024does}.


Super-resolution aims to reconstruct high-frequency components from low-resolution inputs and has been extensively studied in computer vision \cite{wang2020deep, lepcha2023image}. Classical SR methods rely on fixed interpolation techniques such as nearest-neighbor, bilinear, or bicubic upsampling, which are computationally efficient but tend to oversmooth fine details and fail to recover discriminative high-frequency information. With the advent of deep learning, modern SR approaches such as SRCNN~\cite{dong2015image} introduced end-to-end convolutional learning for image reconstruction, while GAN-based methods like SRGAN~\cite{ledig2017photo} emphasized perceptual realism through adversarial and perceptual losses. More recently, transformer-based models such as SwinIR~\cite{liang2021swinir} have achieved state-of-the-art reconstruction performance by leveraging hierarchical attention and long-range dependencies.

Despite their strong visual fidelity, these methods are predominantly optimized for perceptual similarity metrics (e.g., PSNR, SSIM, LPIPS~\cite{zhang2018unreasonable}), which do not necessarily correlate with downstream task performance. Adversarial and perceptual objectives, while effective at improving visual realism, are known to introduce hallucinated high-frequency details that are visually plausible yet task-irrelevant. This phenomenon has been extensively observed in GAN-based super-resolution methods such as SRGAN \cite{ledig2017photo} and ESRGAN \cite{wang2018esrgan}, where adversarial training encourages the synthesis of realistic textures that may not correspond to the underlying image content. This behavior highlights a fundamental perception--distortion trade-off \cite{blau2018perception}, where improvements in perceptual quality may occur at the expense of fidelity to the underlying signal. 

In task-driven settings such as person re-identification, these hallucinated details may degrade downstream performance by introducing semantically misleading cues. This observation has motivated task-oriented super-resolution methods, such as SR4IR~\cite{kim2024beyond}, which explicitly optimize super-resolution for task based objectives rather than pixel-level reconstruction. These findings suggest that blindly applying perceptual or GAN-based super-resolution may be suboptimal for recognition tasks such as person re-identification, where identity preservation and feature consistency are critical.

Despite recent progress, integrating super-resolution (SR) into person re-identification (ReID) remains challenging due to data quality, optimization instability, and task misalignment. Most SR-based ReID methods rely on paired low- and high-resolution supervision, which is rarely available in real-world surveillance systems. Consequently, existing approaches depend heavily on synthetically downsampled data, which do not account for realistic degradations such as motion blur, sensor noise, compression artifacts, and viewpoint variations. This domain gap limits the generalization of SR-ReID models to real deployments.

Many existing approaches rely on GAN-based super-resolution to reconstruct high-frequency details \cite{li2019recover}. Although effective in enhancing perceptual quality, adversarial training is inherently unstable and highly sensitive to hyperparameter choices. More critically, GAN-based methods may hallucinate visually plausible yet identity-inconsistent details, thereby distorting discriminative cues that are essential for reliable person re-identification. In law enforcement applications, such behavior imposes additional constraints, as hallucinated visual evidence can result in incorrect identity associations, ultimately raising concerns about the reliability, interpretability, and evidentiary validity of these approaches.

Super-resolution modules are commonly optimized using pixel-wise reconstruction \cite{dong2015image} or perceptual losses that are agnostic to the ReID objective. As a result, recovered images may preserve global structure while failing to reconstruct fine-grained identity information. Even in joint SR-ReID frameworks, the identity supervision signal is often dominated by reconstruction or adversarial losses, leading to suboptimal task alignment. End-to-end SR-ReID pipelines significantly increase model complexity and training difficulty. Joint optimization can suffer from gradient interference between reconstruction and recognition objectives, resulting in unstable convergence. Additionally, SR networks introduce considerable computational overhead during training and inference, limiting scalability to large-scale datasets and real-time applications.

In summary, while super-resolution can mitigate resolution mismatches in person re-identification, its reliance on paired data, adversarial training, and reconstruction-driven objectives introduces fundamental limitations. These challenges motivate the development of resolution-aware ReID approaches that directly optimize task-relevant representations without explicitly prioritizing image reconstruction. 

In this paper, we introduce S3-CLIP, a GAN-free, task-driven super-resolution framework designed to function under limited paired HR--LR data and significant distribution shifts. Our contributions can be listed as follows:

\begin{itemize}
    \item We present the first video super-resolution framework specifically designed for cross-view person re-identification, addressing extreme resolution mismatches in ground-to-aerial scenarios.
    
    \item We adopt a task-driven two-phase training strategy~\cite{kim2024beyond} that jointly optimizes SwinIR-based super-resolution with CLIP-based ReID objectives, eliminating the need for adversarial training.
    
    \item We propose a temporal consistency loss that enforces smooth super-resolution across video frames, preventing temporal artifacts and maintaining stable identity representations throughout video tracklets.
    
    \item We extensively evaluate our model on the DetReIDX dataset \cite{hambarde2025detreidx}, demonstrating that super-resolution as a pre-processing step achieves 11.24\% rank-1 improvement in ground-to-aerial scenarios, while also analyzing failure modes and limitations.

\end{itemize}


\section{Related works}

In the field of person re-identification, super-resolution has been applied in a variety of ways. Existing SR-based ReID methods can be broadly categorized into image-level reconstruction approaches and representation-level resolution adaptation techniques.

Jiao \emph{et al.}~\cite{jiao2018deep} introduced one of the earliest deep learning frameworks for low-resolution person ReID by jointly optimizing super-resolution reconstruction and identity classification in a unified multi-task learning setup. Their end-to-end design encourages the SR module to preserve identity-discriminative details rather than purely perceptual quality. They also formalized the cross-resolution ReID problem and proposed multi-level resolution benchmarks via synthetic downsampling.

Li \emph{et al.}~\cite{li2019recover} proposed a generative dual-model framework that explicitly separates image recovery and identity recognition. Their approach employs a GAN-based super-resolution network jointly trained with a ReID model under a dual learning scheme, enforcing consistency between recovered images and discriminative identity features.

Adil \emph{et al.}~\cite{adil2020multi} addressed the limitations of fixed-scale super-resolution by introducing a multi-scale adaptive SR framework using GANs. Multiple SR generators trained at different upscaling factors are adaptively fused, enabling robustness to varying resolution degradations commonly observed in real-world ReID scenarios.

Zhang \emph{et al.}~\cite{zhang2021deep} proposed a deep high-resolution representation learning approach that avoids explicit image reconstruction. Their method aligns low- and high-resolution feature distributions by enforcing high-resolution representation constraints, reducing reliance on super-resolved images and mitigating SR-induced artifacts.

Han \emph{et al.}~\cite{han2021adaptive} introduced Adaptive Person Super-Resolution (APSR), which jointly learns multiple SR branches corresponding to different scaling factors. The outputs are adaptively fused to extract complementary identity features, demonstrating that multi-scale SR can balance detail recovery and artifact suppression more effectively than single-scale approaches.

\label{sec:formatting}
\begin{figure*}[ht]
    \centering
    \includegraphics[width=\linewidth]{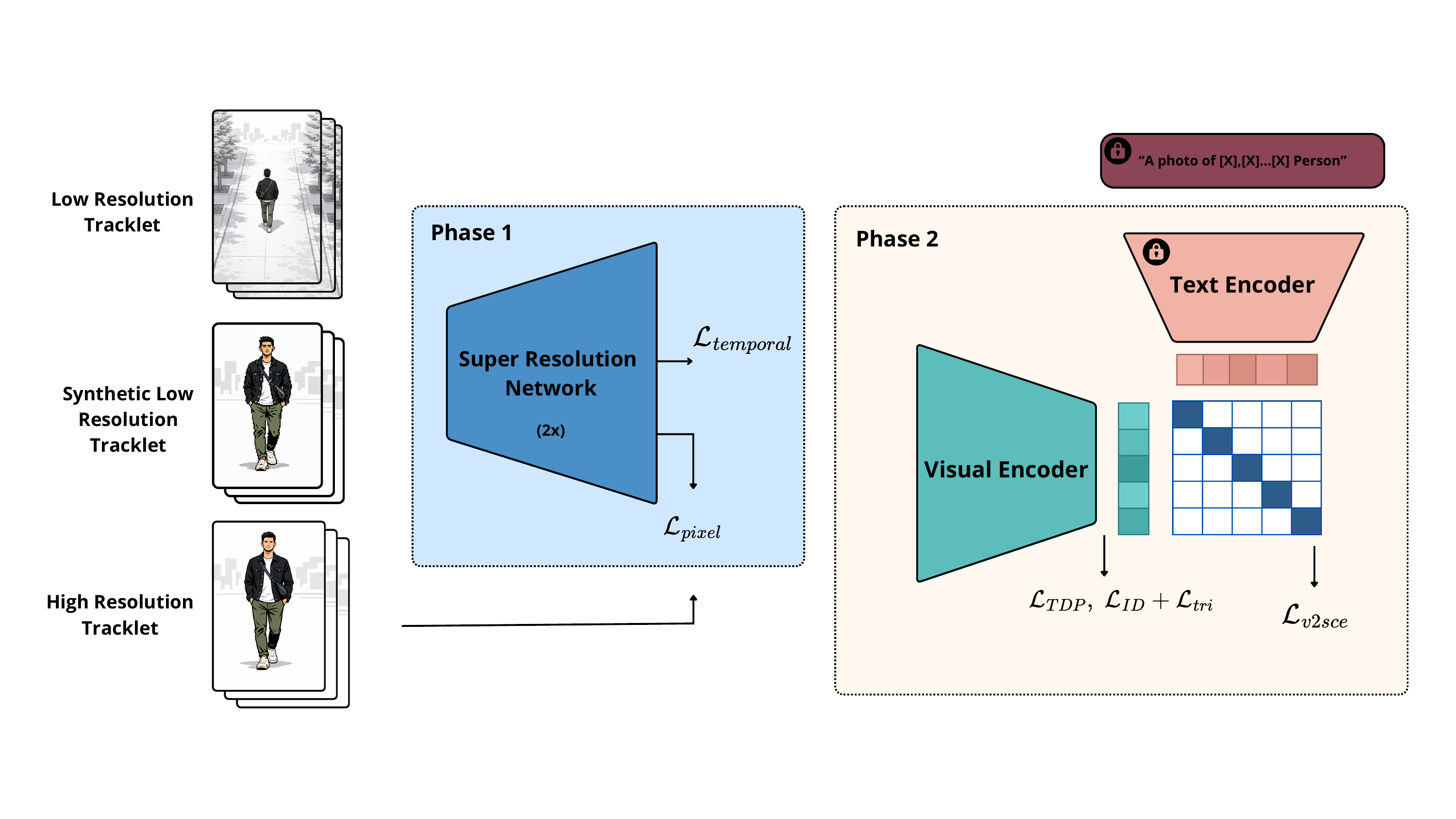}
    \caption{Overview of the S3-CLIP architecture during the second-stage of training. Two types of tracklets are sampled according to SING: (i) a high-resolution tracklet that is synthetically downscaled, and (ii) a naturally low-resolution tracklet of the same person. These low-resolution tracklets are concatenated and fed into the super-resolution network. The outputs are then bicubically upscaled to match the input size of the CLIP-based visual encoder. A two-phase training strategy is employed: first, only the super-resolution network is updated while the visual encoder remains frozen; then, the super-resolution network is frozen and the visual encoder is updated.}
    \label{fig:S3-CLIP}
\end{figure*}

\section{Proposed Method}

 Figure \ref{fig:S3-CLIP} presents an overview of the S3-CLIP architecture. Our method takes inspiration from SING-based sampling \cite{jiao2018deep}, SwinIR upscaler \cite{liang2021swinir}, and the SR4IR two-phase training method \cite{kim2024beyond}. Our method is compatible with any ReID backbone, and the super-resolution module can be easily replaced with alternative upscaling models. In the following subsections, we first describe the baseline ReID pipeline, then detail how super-resolution and the temporal consistency loss are integrated into the framework.

\subsection{Baseline algorithm}

We build upon VSLA-CLIP \cite{zhang2024cross}, a CLIP-ReID~\cite{li2023clip} variant designed for cross-platform visual alignment. VSLA-CLIP adapts the image-based CLIP foundation model to video ReID through a parameter-efficient Video Set-Level-Adapter (VSLA) module, which processes video tracklets while maintaining computational efficiency. The method also introduces platform-bridge prompts to reduce domain discrepancies between aerial and ground camera views. In our framework, VSLA-CLIP serves as the ReID backbone, providing identity-discriminative features for super-resolved video frames. 

\paragraph{CLIP-ReID}

CLIP-ReID \cite{li2023clip} adapts the CLIP vision-language model \cite{radford2021learning} for person re-identification through a two-stage training strategy. This approach achieves strong performance by exploiting CLIP's robust visual features pre-trained on large-scale image-text pairs, which provide better generalization to diverse surveillance scenarios.

The first stage involves the training of a set of ID-specific learnable token parameters \cite{zhou2022learning}, while the visual encoder $\mathcal{I}(\cdot)$ and the text encoder $\mathcal{T}(\cdot)$ remain frozen. The prompts learned here come in the forms of "A photo of a $[X]_1$, $[X]_2$, $[X]_M$ person", where every $[X]_i$, $i \in (1,M)$ is a learnable text token with the same word embedding dimension. $M$ indicates the amount of learnable text tokens. We optimize these tokens using bidirectional contrastive losses: image-to-text $\mathcal{L}_{i2t}$ and text-to-image $\mathcal{L}_{t2i}$. The text-to-image loss for identity $y_i$ is defined as:

\begin{equation}
    \mathcal{L}_{t2i}(y_i)
=
-\frac{1}{\lvert P(y_i) \rvert}
\sum_{p \in P(y_i)}
\log
\frac{
\exp\!\left( s(\mathbf{I}_p, \mathcal{T}(y_i)) \right)
}{
\sum_{a=1}^{B}
\exp\!\left( s(\mathbf{I}_a, \mathcal{T}(y_i)) \right)
}
    \label{eq:t2i}
\end{equation} 

where $P(y_i)$ denotes the set of all image samples belonging to identity $y_i$ within the batch, $\mathbf{I}_p$ is an image from this set, $s(\cdot, \cdot)$ computes the cosine similarity between image and text embeddings, $B$ is the batch size. This loss encourages the text prompt for identity $y_i$ to have high similarity with all images of that identity while maintaining low similarity with images from other identities in the batch. The image-to-text loss $\mathcal{L}_{i2t}$ follows a symmetric formulation, where each image embedding is encouraged to match its corresponding identity's text prompt. The overall loss function of stage 1 $\mathcal{L}_{stage1}$ can be described as their combination in Eq. \ref{eq:stage1}:

\begin{equation}
    \mathcal{L}_{stage1} = \mathcal{L}_{i2t} + \mathcal{L}_{t2i}
    \label{eq:stage1}
\end{equation}

\paragraph{Video Set-Level-Adapter CLIP}

Cross-platform video person re-identification (ReID) faces significant visual domain gaps between camera platforms, such as ground and aerial views. The VSLA-CLIP formulates this problem as a visual-semantic alignment task using a pretrained CLIP model. Learnable ID-specific description tokens $[S]_i$ and shared prompt tokens $[P]_i$ are concatenated and encoded via CLIP's text encoder $\mathcal{T}(\cdot)$ to produce semantic embeddings:

\begin{equation}
\mathbf{T} = \mathcal{T}\Big([\,[P]_1 \dots [P]_{n/2} : [S]_1 \dots [S]_M : [P]_{n/2+1} \dots [P]_n\,]\Big)
\label{eq:text_embedding}
\end{equation}
where $[:\,]$ denotes token concatenation.

Video features are obtained by encoding each frame $V_{ij}$ of a video $V_i$ using CLIP's image encoder $\mathcal{I}(\cdot)$ and aggregating via mean pooling:

\begin{equation}
\mathbf{V}_i = \frac{1}{T}\sum_{j=1}^T \mathcal{I}(\mathcal{V}_{ij})
\label{eq:video_feature}
\end{equation}

with $T$ frames. Alignment between video and text embeddings is enforced through contrastive video-to-semantic loss:

\begin{equation}
\mathcal{L}_{v2sce}(i)=\sum_{k=1}^N -q_k\log\frac{\exp(s(\mathbf{V}_i,\mathbf{T}_{y_k}))}{\sum_{j=1}^N\exp(s(\mathbf{V}_i,\mathbf{T}_{y_j}))},
\label{eq:contrastive_loss}
\end{equation}

where $s(\cdot,\cdot)$ is cosine similarity and $q_k$ soft labels, which labels are determined by online label smoothing \cite{zhang2021delving}. Additional triplet and identity losses are applied:

\begin{equation}
\mathcal{L}_{\text{tri}}=\max(d_p-d_n+\theta,0),\quad 
\mathcal{L}_{ID}=\sum_{k=1}^N -q_k\log(p_k),
\label{eq:tri_id_loss}
\end{equation}
leading to the overall re-identification loss:
\begin{equation}
\mathcal{L}_{ReID}=\mathcal{L}_{v2sce}+\beta\,\mathcal{L}_{\text{tri}}+\gamma\,\mathcal{L}_{ID}+\delta\,\mathcal{L}_{i2t}+\epsilon\,\mathcal{L}_{t2i},
\label{eq:stage2_loss}
\end{equation}
where $\beta,\gamma,\delta,\epsilon$ balance the individual sub-loss contributions. To reduce fine-tuning cost while maintaining temporal context, \emph{Video Set-Level Adapter} (VSLA) is utilized, which includes \emph{Intra-Frame Adapter} (IFA) and \emph{Cross-Frame Attention Adapter} (CFAA). IFA projects activations down and up within each MLP block:
\begin{equation}
\begin{gathered}
\textsc{IFA}(\mathbf{x}') 
= \sigma(\mathbf{x}' W_{\text{down}}) W_{\text{up}}, \\
\mathbf{x}_i 
= \textsc{MLP}(\textsc{LN}(\mathbf{x}_i')) 
 + \mathbf{x}_i' 
 + \textsc{IFA}(\mathbf{x}_i'),
\end{gathered}
\label{eq:ifa_update}
\end{equation}

while CFAA aggregates frame context invariant to frame order:

\begin{equation}
\begin{gathered}
\mathbf{M}(\{\mathcal{V}_{ij}\}) 
= \mathbf{M}(\{\mathcal{V}_{i\pi(j)}\}), \\
\mathbf{x}_i' 
= \textsc{MSA}(\textsc{LN}(\mathbf{x}_{i-1})) 
 + \mathbf{x}_{i-1} 
 + \textsc{CFAA}(\mathbf{x}_{i-1}),
\end{gathered}
\label{eq:cfaa_update}
\end{equation}

where $\pi$ is a permutation and MSA is multi-head self-attention.

Finally, \emph{Platform-Bridge Prompts} (PBP) are integrated into the text stream to encode platform-specific context. This mechanism ensures that visual embeddings from different platforms are semantically aligned, enhancing cross-platform ReID performance without introducing additional explicit losses. In summary, this baseline combines visual-semantic alignment (Eq.~\ref{eq:text_embedding}-\ref{eq:stage2_loss}), parameter-efficient video adaptation via VSLA (Eq.~\ref{eq:ifa_update}-\ref{eq:cfaa_update}), and platform-aware prompts, providing a robust reference framework for cross-platform video person ReID.

\subsection{Sampling for Super Resolution ReID Training}

Real-world surveillance datasets often contain images with widely varying resolutions across different camera views, which makes it challenging to find paired high- and low-quality training samples. To address this limitation, we adopt a semi-supervised training strategy inspired by SING~\cite{jiao2018deep}, adapted to handle unpaired cross-resolution video tracklets.

We decouple the resolution notions of high resolution frames from the input resolution of the CLIP visual encoder. By intentionally setting the threshold $H_h \times W_h$ for high resolution frames lower than the input resolution of the ViT~\cite{dosovitskiy2020image} visual encoder, we maximize the availability of high-low resolution training pairs, while not reducing the amount of patch tokens available. We partition the dataset based on native frame resolution into low-resolution $X^l = \{ (x^l_i, y_i) \}_{i=1}^{N_l}$ and high-resolution $X^h = \{ (x^h_i, y_i) \}_{i=1}^{N_h}$ tracklet sets.

During training, we sample triplets for each batch entry. First, we select a high-resolution tracklet from $X^h$ whose resolution exceeds a threshold $H_h \times W_h$. This ensures that a diverse set of tracklets is included with sufficient high-frequency details. These high-resolution tracklets are then bicubically upsampled to match the input size of the ReID encoder. 

Next, we generate synthetic low-resolution tracklets $X^{h2l}$ from $X^h$ and sample naturally low-resolution tracklets. These tracklets are resized to a uniform resolution of $D \cdot (H_h \times W_h)$ and concatenated. This procedure results in a semi-supervised batch containing $P$ identities with $K$ triplets each, with each triplet containing a high resolution tracklet, synthetic low resolution tracklet and a natural low resolution tracklet.

\subsection{Video Super Resolution Loss}

For the super-resolution component of S3-CLIP, we employ SwinIR~\cite{liang2021swinir} as our super-resolution model, denoted by $\mathcal{S}(\cdot)$. During training, we optimize a combination of losses. First, the pixel loss $\mathcal{L}_{pixel}$ measures the L1 norm between high-resolution tracklet frames $x_t^{h}$ and their super-resolved counterparts from synthetically downsampled inputs $x_t^{h2l}$. In addition, we use a task-driven perceptual loss $\mathcal{L}_{TDP}$ (Eq.~\ref{eq:TDP loss})~\cite{kim2024beyond}, computed only from the last layer of the visual encoder, before the BNN neck. To ensure temporal consistency within tracklets, we also employ a temporal loss $\mathcal{L}_{temporal}$ (Eq.~\ref{eq:temporal_loss}).

\begin{equation}
    \mathcal{L}_{TDP} = || \mathcal{I}(X^h) - \mathcal{I}(\mathcal{S}(X^{h2l}))||_1
    \label{eq:TDP loss}
\end{equation}

\begin{equation}
\begin{aligned}
\mathcal{L}_{temporal}
&= \frac{1}{T-1} \sum_{t=1}^{T-1}
\Big\|
\big( \mathcal{S}(x^{h2l}_{t+1}) - \mathcal{S}(x^{h2l}_t) \big) \\
&\quad - \big( x^{h}_{t+1} - x^{h}_t \big)
\Big\|_1
\end{aligned}
\label{eq:temporal_loss}
\end{equation}

For training the super-resolution module in the ReID setting, we adopt the two-phase optimization strategy introduced in SR4IR~\cite{kim2024beyond}. Jointly optimizing super-resolution and ReID networks has been shown to cause gradient conflicts, which can negatively impact performance~\cite{kim2024beyond}. To address this issue, SR4IR decomposes the joint optimization into two sequential phases.

In the first phase, the visual encoder is frozen and only the super-resolution network is updated by minimizing the super-resolution loss $\mathcal{L}_{SR} = \mathcal{L}_{pixel} + \mathcal{L}_{TDP} + \mathcal{L}_{temporal}$. In the second phase, the super-resolution network is frozen, and the ReID visual encoder is updated using the ReID objectives $\mathcal{L}_{ReID}$.

Let $\boldsymbol{y}$ denote the identity labels of the sampled tracklets, and $\theta_{\mathrm{ReID}}$ and $\theta_{\mathrm{SR}}$ be the parameters of the visual encoder $\mathcal{I}(\cdot)$ and the super-resolution network $\mathcal{S}(\cdot)$, respectively. The second-stage training objective is defined as Eq. \ref{eq:stage2}:

\begin{equation}
\begin{aligned}
\min_{\boldsymbol{\theta}_{\text{SR}}} \quad & \mathcal{L}_{\text{SR}}(S_{\boldsymbol{\theta}_{\text{SR}}}(X^l,X^{h2l}), X^h), & \text{in phase 1,} \\
\min_{\boldsymbol{\theta}_{\text{ReID}}} \quad & \mathcal{L}_{\text{ReID}}(\mathcal{I}_{\boldsymbol{\theta}_{\text{ReID}}}(X^l,X^{h2l}, X^h), \boldsymbol{y}), & \text{in phase 2,}
\end{aligned}
\label{eq:stage2}
\end{equation}

\section{Experiments}

\paragraph{Dataset.} We evaluate our method on the DetReIDX~\cite{hambarde2025detreidx} video-based person re-identification dataset. DetReIDX simulates challenging real-world surveillance conditions with substantial variations in resolution, viewpoint, scale, occlusion, and appearance across aerial and ground-based camera systems. The dataset comprises over 13 million bounding boxes spanning 509 identities across seven sites.

\paragraph{Evaluation protocol.} We evaluate our method under three query--gallery matching settings: aerial-to-aerial ($A \rightarrow A$), aerial-to-ground ($A \rightarrow G$), and ground-to-aerial ($G \rightarrow A$). Re-identification performance is assessed using Rank-1, Rank-5, Rank-10 accuracy and mean Average Precision (mAP). For further analysis, we compare two optimization strategies for S3-CLIP: joint optimization of the super-resolution and ReID networks ($S3$-CLIP: $\mathcal{S} + \mathcal{I}$) and the proposed two-phase optimization ($S3$-CLIP). As a baseline, we employ VSLA-CLIP with bilinear upsampling.

\begin{table*}[ht]
\centering
\resizebox{\textwidth}{!}{%
\begin{tabular}{l|cccc|cccc|cccc|c}
\toprule
\multirow{2}{*}{\textbf{Method}} & \multicolumn{4}{c|}{\textbf{A→A}} & \multicolumn{4}{c|}{\textbf{A→G}} & \multicolumn{4}{c|}{\textbf{G→A}} & \multirow{2}{*}{\textbf{Overall mAP}} \\
\cmidrule{2-13}
 & R1 & R5 & R10 & mAP & R1 & R5 & R10 & mAP & R1 & R5 & R10 & mAP & \\
\midrule
\multicolumn{14}{l}{\textbf{Baseline Method}} \\
\midrule
VSLA-CLIP~\cite{zhang2024cross} & \textbf{18.75} & 28.22 & 35.03 & \textbf{15.99} & 31.21 & \textbf{58.83} & 73.88 & 37.87 & 57.30 & 62.92 & 66.29 & 27.45 & \textbf{27.69} \\
\midrule
\multicolumn{14}{l}{\textbf{Proposed Method}} \\
\midrule
$S3-CLIP: \mathcal{S} + \mathcal{I}$ & 18.68 & \textbf{30.84} & \textbf{39.55} & 14.83 & \textbf{31.90} & 58.71 & \textbf{75.00} & \textbf{38.36} & 64.04 & 71.91 & 76.40 & 29.00 & 27.44 \\
$S3-CLIP$ & 17.43 & 30.30 & 39.15 & 14.24 & 31.09 & 57.18 & 72.75 & 37.52 & \textbf{68.54} & \textbf{76.40} & \textbf{84.27} & \textbf{29.16} & 26.72 \\
\bottomrule
\end{tabular}%
}
\caption{Comparison of proposed methods and baseline on DetReIDX. Best results are highlighted in bold per protocol and metric.}
\label{tab:vreid_xfd_results}
\end{table*}

\begin{figure*}[t] \centering \begin{subfigure}{0.32\textwidth} \centering \includegraphics[width=0.32\textwidth]{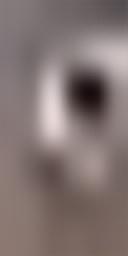} \caption{Extreme low resolution.} \end{subfigure} \begin{subfigure}{0.32\textwidth} \centering \includegraphics[width=0.32\textwidth]{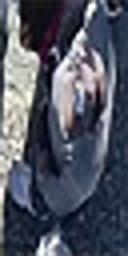} \caption{Aspect ratio mismatch} \end{subfigure} \begin{subfigure}{0.32\textwidth} \centering \includegraphics[width=0.32\textwidth]{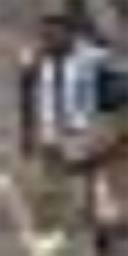} \caption{Motion Blur and JPG compression.} \end{subfigure} \caption{Representative failure cases where super-resolution degrades ReID performance due to extremely low resolution, aspect ratio mismatch, JPG compression and motion blur.} \label{fig:fail_cases} \end{figure*}

\paragraph{Network Architecture.} For the SR network architecture, we employ SwinIR-S~\cite{liang2021swinir}, an 910k parameter variant, which was pre-trained on the DIV2K \cite{agustsson2017ntire} dataset. For the visual encoder VIT-B-16 \cite{dosovitskiy2020image} is used. The networks were implemented in Pytorch \cite{paszke2019pytorch} and ImageNet \cite{deng2009imagenet} normalization was used.

\paragraph{Implementation details.} We use the Adam optimizer~\cite{kingma2017adam} for both the super-resolution and ReID networks. The learning rate in the first stage is set to $3.5 \times 10^{-4}$. In the second stage, the learning rate is $7.5 \times 10^{-5}$ for the visual encoder and $7.5 \times 10^{-6}$ for the super-resolution network.

The normalization strategy differs between variants. The joint optimization variant ($S3$-CLIP: $\mathcal{S} + \mathcal{I}$) applies ImageNet normalization before the super resolution network, using its output for the visual encoder, whereas $S3$-CLIP retains image values in the $[0,1]$ range for the super resolution network and applies ImageNet normalization before the visual encoder. The weight decay in the first stage is set to $10^{-4}$, while in the second stage both the visual encoder and the super-resolution network use a weight decay of $5 \times 10^{-3}$.

For stage two, we employ the same learning rate scheduler for both networks, consisting of a 10-epoch warmup with a warmup factor of $0.1$. The batch size is set to 16 in the first stage and 8 in the second stage, with gradient accumulation of 3 during the second stage. Each batch contains 8 frames per tracklet and 4 distinct identities, with all frames resized to $256 \times 128$.

For S3-CLIP sampling, frames with resolution above $H_h \times W_h = 128 \times 64$ are treated as high resolution and upsampled using bicubic interpolation. Frames below this threshold are resized to a uniform resolution of $D \cdot (H_h \times W_h) = 64 \times 32$, where $D = \{\frac{1}{2}\}$, then upscaled using a $2\times$ SwinIR model, and finally bicubically upsampled to the input resolution. The margin for the triplet loss is set to $0.3$, and the ReID loss $\mathcal{L}_{ReID}$ weighting coefficients $\beta$, $\gamma$, $\delta$, and $\epsilon$ are set to $1.0$, $0.25$, $1.0$, and $1.0$, respectively. Data augmentation includes horizontal flipping and consistent random erasing~\cite{zhong2020random}, applied uniformly across all frames within a tracklet.

\subsection{Quantitative Results}

Table \ref{tab:vreid_xfd_results} presents the quantitative person ReID performance on the DetReIDX benchmark across three matching protocols. Our proposed methods demonstrate protocol-specific improvements over the baseline VSLA-CLIP approach. In the aerial-to-ground (A→G) scenario, the jointly optimized $S3-CLIP: \mathcal{S} + \mathcal{I}$ variant achieves the best results, improving R1 by $0.69\%$, R10 by $1.12\%$, and mAP by $0.49\%$. For the ground-to-aerial (G→A) protocol, both variants substantially outperform the baseline, with $S3-CLIP$ achieving improvements of $11.24\%$ in R1, $13.48\%$ in R5, $17.98\%$ in R10, and $1.71\%$ in mAP. The $\mathcal{S} + \mathcal{I}$ variant shows gains of $6.74\%$, $9.00\%$, $10.11\%$, and $1.55\%$ for the same metrics, respectively. In the aerial-to-aerial (A→A) matching scenario, both methods achieve competitive performance, with $\mathcal{S} + \mathcal{I}$ improving R5 and R10 by $2.62\%$ and $4.52\%$ while maintaining comparable mAP ($-1.16\%$). These results indicate that super-resolution pre-processing provides the most substantial benefits for cross-view matching scenarios, particularly when matching ground-level queries to aerial galleries.






\subsection{Fail cases}

To better understand the limitations of our method, we visualize representative failure cases in Figure~\ref{fig:fail_cases}. These failures arise mainly in scenarios that involve extremely low-resolution inputs, underrepresented training conditions, and strong image degradations, including motion blur and JPEG compression.

As input resolution decreases, the available information for reconstruction diminishes correspondingly. At extreme low resolutions such as $6 \times 6$ pixels present in DetReIDX~\cite{hambarde2025detreidx} , it becomes extraordinarily difficult to recover or even generate~\cite{chen2024deep} higher-resolution images that preserve identity-relevant details. Therefore we mostly see in such cases that SwinIR~\cite{liang2021swinir} extrapolates the blurry artifacts instead of hallucinating high-quality details.

Super-resolution alone is insufficient to recover information lost due to motion blur or strong JPEG compression~\cite{kim2025exploiting}. These degradations are not strictly resolution-related and often remove or corrupt high-frequency identity cues in a non-invertible manner. As a result, SR may amplify compression artifacts or blurred background structures, which can negatively impact downstream ReID performance. Our method, trained primarily on bicubic downsampling, struggles to generalize to these authentic degradation patterns.

The aspect ratio mismatch is inherent even in our algorithm, since we only deal with improving the resolution of the tracklets. ReID pipelines, most commonly resize images to a fixed input size for the network, which can distort the original aspect ratio of the person. Distorted shapes change the geometric relationships of body parts and can shift the distribution of learned features, making identity features harder for the network to reliably encode.


\section{Discussion}

Experimental results demonstrate that super-resolution pre-processing provides substantial improvements for cross-view person re-identification, particularly in ground-to-aerial matching scenarios where query resolution is significantly lower than gallery resolution. However, several factors limit the generalization and scalability of our approach.

While our semi-supervised sampling strategy mitigates the lack of paired high- and low-resolution data, residual domain gaps persist across camera views. Certain cameras remain underrepresented in the synthetic HR-to-LR training distribution, leading to imbalanced supervision that disproportionately affects aerial scenarios where native resolution is lower. Furthermore, naturally low-resolution tracklets without corresponding high-resolution counterparts receive only identity supervision without explicit reconstruction guidance, limiting the SR network's ability to generalize to authentic degradations.

We observe substantial asymmetry between aerial-to-ground ($A\!\rightarrow\!G$) and ground-to-aerial ($G\!\rightarrow\!A$) matching performance. The $G\!\rightarrow\!A$ protocol contains only 89 query tracklets compared to 5,682 in $A\!\rightarrow\!G$, making retrieval highly sensitive to gallery coverage and resolution variance. This explains the larger relative improvements ($+11.24\%$ R1) in $G\!\rightarrow\!A$ versus $A\!\rightarrow\!G$ ($+0.69\%$ R1). The aerial-to-aerial ($A\!\rightarrow\!A$) setting remains particularly challenging due to extreme scale variation and limited discriminative detail at both query and gallery levels, where super-resolution so far provides minimal benefit.

Our single-scale super-resolution model with fixed $2\times$ upsampling limits adaptability to diverse resolution distributions. While controlled experiments demonstrate this design sufficiently improves baseline performance~\cite{jiao2018deep}, real-world deployments exhibit more complex degradation patterns including motion blur, JPEG compression, and sensor noise (Figure~\ref{fig:fail_cases}). Additionally, bicubic pre-interpolation before super-resolution may introduce aliasing artifacts.

Increasing input resolution or the number of frames per tracklet consistently improves performance~\cite{nguyen2025ag}, but incurs significant memory and computational overhead. This trade-off is particularly relevant for large-scale benchmarks such as DetReIDX, where long tracklets and high-resolution inputs can quickly exceed practical resource constraints.

\section{Conclusion}

In this paper, we present S3-CLIP, the first video super-resolution framework for person re-identification. By integrating SwinIR-based super-resolution with VSLA-CLIP through a task-driven two-phase training strategy and temporal consistency loss, we demonstrate that super-resolution can serve as an effective preprocessing step for video person ReID, particularly in challenging ground-to-aerial scenarios.

Our experiments on DetReIDX reveal protocol-specific benefits: S3-CLIP achieves substantial improvements in ground-to-aerial matching (+11.24\% R1, +17.98\% R10), while maintaining competitive performance in aerial-to-ground scenarios. These results validate that task-driven super-resolution, can recover identity-discriminative details critical for cross-resolution matching without relying on adversarial training.

Through comprehensive failure case analysis, we identify fundamental limitations: extreme low-resolution inputs, authentic degradations (motion blur, JPEG compression), and aspect ratio distortions remain challenging for current super-resolution approaches. These findings highlight important directions for future work, including multi-scale adaptive super-resolution, authentic degradation modeling.

As visual encoder architectures continue to evolve, our backbone-agnostic design ensures that S3-CLIP can be readily integrated into existing ReID pipelines, providing a practical and scalable solution for real-world surveillance applications where resolution mismatch is unavoidable.

\section*{Acknowledgments}

This work was supported in part by Phase Transition Materials for Energy Efficient Edge Computing (PHASTRAC, G.A. No. 101092096), the Hungarian Government Thematic Excellence Programme (TKP2021-NVA-27), and the Hungarian Ministry of Culture and Innovation through the National Research, Development and Innovation Office (Project No. 2022-1.2.5-TÉT-IPARI-KR-2022-00015).

{
    \small
    \bibliographystyle{ieeenat_fullname}
    \bibliography{main}

@String(ECCV= {Eur. Conf. Comput. Vis.})

@String(AAAI = {AAAI})

@String(ECCV  = {ECCV})

@inproceedings{zhang2023ground,
  title={Ground-to-aerial person search: Benchmark dataset and approach},
  author={Zhang, Shizhou and Yang, Qingchun and Cheng, De and Xing, Yinghui and Liang, Guoqiang and Wang, Peng and Zhang, Yanning},
  booktitle={Proceedings of the 31st ACM International Conference on Multimedia},
  pages={789--799},
  year={2023}
}

@inproceedings{nguyen2025ag,
  title={AG-VPReID: A Challenging Large-Scale Benchmark for Aerial-Ground Video-based Person Re-Identification},
  author={Nguyen, Huy and Nguyen, Kien and Pemasiri, Akila and Liu, Feng and Sridharan, Sridha and Fookes, Clinton},
  booktitle={Proceedings of the Computer Vision and Pattern Recognition Conference},
  pages={1241--1251},
  year={2025}
}

@article{hambarde2025detreidx,
  title={DetReIDX: A stress-test dataset for real-world UAV-based person recognition},
  author={Hambarde, Kailash A and Mbongo, Nzakiese and MP, Pavan Kumar and Mekewad, Satish and Fernandes, Carolina and Silahtaro{\u{g}}lu, G{\"o}khan and Nithya, Alice and Wasnik, Pawan and Rashidunnabi, MD and Samale, Pranita and others},
  journal={arXiv preprint arXiv:2505.04793},
  year={2025}
}

@inproceedings{cheng2020inter,
  title={Inter-task association critic for cross-resolution person re-identification},
  author={Cheng, Zhiyi and Dong, Qi and Gong, Shaogang and Zhu, Xiatian},
  booktitle={Proceedings of the IEEE/CVF conference on computer vision and pattern recognition},
  pages={2605--2615},
  year={2020}
}

@inproceedings{radford2021learning,
  title={Learning transferable visual models from natural language supervision},
  author={Radford, Alec and Kim, Jong Wook and Hallacy, Chris and Ramesh, Aditya and Goh, Gabriel and Agarwal, Sandhini and Sastry, Girish and Askell, Amanda and Mishkin, Pamela and Clark, Jack and others},
  booktitle={International conference on machine learning},
  pages={8748--8763},
  year={2021},
  organization={PmLR}
}

@inproceedings{caron2021emerging,
  title={Emerging properties in self-supervised vision transformers},
  author={Caron, Mathilde and Touvron, Hugo and Misra, Ishan and J{\'e}gou, Herv{\'e} and Mairal, Julien and Bojanowski, Piotr and Joulin, Armand},
  booktitle={Proceedings of the IEEE/CVF international conference on computer vision},
  pages={9650--9660},
  year={2021}
}

@article{oquab2023dinov2,
  title={Dinov2: Learning robust visual features without supervision},
  author={Oquab, Maxime and Darcet, Timoth{\'e}e and Moutakanni, Th{\'e}o and Vo, Huy and Szafraniec, Marc and Khalidov, Vasil and Fernandez, Pierre and Haziza, Daniel and Massa, Francisco and El-Nouby, Alaaeldin and others},
  journal={arXiv preprint arXiv:2304.07193},
  year={2023}
}

@inproceedings{jiao2018deep,
  title={Deep low-resolution person re-identification},
  author={Jiao, Jiening and Zheng, Wei-Shi and Wu, Ancong and Zhu, Xiatian and Gong, Shaogang},
  booktitle={Proceedings of the AAAI conference on artificial intelligence},
  volume={32},
  number={1},
  year={2018}
}

@article{zhang2021deep,
  title={Deep high-resolution representation learning for cross-resolution person re-identification},
  author={Zhang, Guoqing and Ge, Yu and Dong, Zhicheng and Wang, Hao and Zheng, Yuhui and Chen, Shengyong},
  journal={IEEE Transactions on Image processing},
  volume={30},
  pages={8913--8925},
  year={2021},
  publisher={IEEE}
}

@article{adil2020multi,
  title={Multi scale-adaptive super-resolution person re-identification using GAN},
  author={Adil, Muhammad and Mamoon, Saqib and Zakir, Ali and Manzoor, Muhammad Arslan and Lian, Zhichao},
  journal={Ieee Access},
  volume={8},
  pages={177351--177362},
  year={2020},
  publisher={IEEE}
}

@article{han2021adaptive,
  title={Adaptive super-resolution for person re-identification with low-resolution images},
  author={Han, Ke and Huang, Yan and Song, Chunfeng and Wang, Liang and Tan, Tieniu},
  journal={Pattern Recognition},
  volume={114},
  pages={107682},
  year={2021},
  publisher={Elsevier}
}

@inproceedings{li2019recover,
  title={Recover and identify: A generative dual model for cross-resolution person re-identification},
  author={Li, Yu-Jhe and Chen, Yun-Chun and Lin, Yen-Yu and Du, Xiaofei and Wang, Yu-Chiang Frank},
  booktitle={Proceedings of the IEEE/CVF international conference on computer vision},
  pages={8090--8099},
  year={2019}
}

@inproceedings{khalid2025bridging,
  title={Bridging the Sky and Ground: Towards View-Invariant Feature Learning for Aerial-Ground Person Re-Identification},
  author={Khalid, Wajahat and Liu, Bin and Li, Xulin and Waqas, Muhammad and Afgan, Muhammad Sher},
  booktitle={Proceedings of the IEEE/CVF International Conference on Computer Vision},
  pages={9749--9758},
  year={2025}
}

@inproceedings{chen2023beyond,
  title={Beyond appearance: a semantic controllable self-supervised learning framework for human-centric visual tasks},
  author={Chen, Weihua and Xu, Xianzhe and Jia, Jian and Luo, Hao and Wang, Yaohua and Wang, Fan and Jin, Rong and Sun, Xiuyu},
  booktitle={Proceedings of the IEEE/CVF conference on computer vision and pattern recognition},
  pages={15050--15061},
  year={2023}
}

@inproceedings{yu2024tf,
  title={Tf-clip: Learning text-free clip for video-based person re-identification},
  author={Yu, Chenyang and Liu, Xuehu and Wang, Yingquan and Zhang, Pingping and Lu, Huchuan},
  booktitle={Proceedings of the AAAI conference on artificial intelligence},
  volume={38},
  number={7},
  pages={6764--6772},
  year={2024}
}

@article{wang2020deep,
  title={Deep learning for image super-resolution: A survey},
  author={Wang, Zhihao and Chen, Jian and Hoi, Steven CH},
  journal={IEEE transactions on pattern analysis and machine intelligence},
  volume={43},
  number={10},
  pages={3365--3387},
  year={2020},
  publisher={IEEE}
}

@article{lepcha2023image,
  title={Image super-resolution: A comprehensive review, recent trends, challenges and applications},
  author={Lepcha, Dawa Chyophel and Goyal, Bhawna and Dogra, Ayush and Goyal, Vishal},
  journal={Information Fusion},
  volume={91},
  pages={230--260},
  year={2023},
  publisher={Elsevier}
}

@article{dong2015image,
  title={Image super-resolution using deep convolutional networks},
  author={Dong, Chao and Loy, Chen Change and He, Kaiming and Tang, Xiaoou},
  journal={IEEE transactions on pattern analysis and machine intelligence},
  volume={38},
  number={2},
  pages={295--307},
  year={2015},
  publisher={IEEE}
}

@inproceedings{ledig2017photo,
  title={Photo-realistic single image super-resolution using a generative adversarial network},
  author={Ledig, Christian and Theis, Lucas and Husz{\'a}r, Ferenc and Caballero, Jose and Cunningham, Andrew and Acosta, Alejandro and Aitken, Andrew and Tejani, Alykhan and Totz, Johannes and Wang, Zehan and others},
  booktitle={Proceedings of the IEEE conference on computer vision and pattern recognition},
  pages={4681--4690},
  year={2017}
}

@inproceedings{liang2021swinir,
  title={Swinir: Image restoration using swin transformer},
  author={Liang, Jingyun and Cao, Jiezhang and Sun, Guolei and Zhang, Kai and Van Gool, Luc and Timofte, Radu},
  booktitle={Proceedings of the IEEE/CVF international conference on computer vision},
  pages={1833--1844},
  year={2021}
}

@inproceedings{zhang2018unreasonable,
  title={The unreasonable effectiveness of deep features as a perceptual metric},
  author={Zhang, Richard and Isola, Phillip and Efros, Alexei A and Shechtman, Eli and Wang, Oliver},
  booktitle={Proceedings of the IEEE conference on computer vision and pattern recognition},
  pages={586--595},
  year={2018}
}

@article{sundaram2024does,
  title={When does perceptual alignment benefit vision representations?},
  author={Sundaram, Shobhita and Fu, Stephanie and Muttenthaler, Lukas and Tamir, Netanel and Chai, Lucy and Kornblith, Simon and Darrell, Trevor and Isola, Phillip},
  journal={Advances in Neural Information Processing Systems},
  volume={37},
  pages={55314--55341},
  year={2024}
}

@inproceedings{wang2018esrgan,
  title={Esrgan: Enhanced super-resolution generative adversarial networks},
  author={Wang, Xintao and Yu, Ke and Wu, Shixiang and Gu, Jinjin and Liu, Yihao and Dong, Chao and Qiao, Yu and Change Loy, Chen},
  booktitle={Proceedings of the European conference on computer vision (ECCV) workshops},
  pages={0--0},
  year={2018}
}

@inproceedings{blau2018perception,
  title={The perception-distortion tradeoff},
  author={Blau, Yochai and Michaeli, Tomer},
  booktitle={Proceedings of the IEEE conference on computer vision and pattern recognition},
  pages={6228--6237},
  year={2018}
}

@inproceedings{zhang2024cross,
  title={Cross-platform video person reid: A new benchmark dataset and adaptation approach},
  author={Zhang, Shizhou and Luo, Wenlong and Cheng, De and Yang, Qingchun and Ran, Lingyan and Xing, Yinghui and Zhang, Yanning},
  booktitle={European Conference on Computer Vision},
  pages={270--287},
  year={2024},
  organization={Springer}
}

@inproceedings{li2023clip,
  title={Clip-reid: exploiting vision-language model for image re-identification without concrete text labels},
  author={Li, Siyuan and Sun, Li and Li, Qingli},
  booktitle={Proceedings of the AAAI conference on artificial intelligence},
  volume={37},
  number={1},
  pages={1405--1413},
  year={2023}
}

@article{zhou2022learning,
  title={Learning to prompt for vision-language models},
  author={Zhou, Kaiyang and Yang, Jingkang and Loy, Chen Change and Liu, Ziwei},
  journal={International Journal of Computer Vision},
  volume={130},
  number={9},
  pages={2337--2348},
  year={2022},
  publisher={Springer}
}

@inproceedings{kim2024beyond,
  title={Beyond image super-resolution for image recognition with task-driven perceptual loss},
  author={Kim, Jaeha and Oh, Junghun and Lee, Kyoung Mu},
  booktitle={Proceedings of the IEEE/CVF Conference on Computer Vision and Pattern Recognition},
  pages={2651--2661},
  year={2024}
}

@inproceedings{agustsson2017ntire,
  title={Ntire 2017 challenge on single image super-resolution: Dataset and study},
  author={Agustsson, Eirikur and Timofte, Radu},
  booktitle={Proceedings of the IEEE conference on computer vision and pattern recognition workshops},
  pages={126--135},
  year={2017}
}

@article{dosovitskiy2020image,
  title={An image is worth 16x16 words: Transformers for image recognition at scale},
  author={Dosovitskiy, Alexey},
  journal={arXiv preprint arXiv:2010.11929},
  year={2020}
}

@article{paszke2019pytorch,
  title={Pytorch: An imperative style, high-performance deep learning library},
  author={Paszke, Adam and Gross, Sam and Massa, Francisco and Lerer, Adam and Bradbury, James and Chanan, Gregory and Killeen, Trevor and Lin, Zeming and Gimelshein, Natalia and Antiga, Luca and others},
  journal={Advances in neural information processing systems},
  volume={32},
  year={2019}
}

@inproceedings{deng2009imagenet,
  title={Imagenet: A large-scale hierarchical image database},
  author={Deng, Jia and Dong, Wei and Socher, Richard and Li, Li-Jia and Li, Kai and Fei-Fei, Li},
  booktitle={2009 IEEE conference on computer vision and pattern recognition},
  pages={248--255},
  year={2009},
  organization={Ieee}
}

@misc{kingma2017adam,
      title={Adam: A Method for Stochastic Optimization}, 
      author={Diederik P. Kingma and Jimmy Ba},
      year={2017},
      eprint={1412.6980},
      archivePrefix={arXiv},
      primaryClass={cs.LG},
      url={https://arxiv.org/abs/1412.6980}, 
}

@inproceedings{zhong2020random,
  title={Random erasing data augmentation},
  author={Zhong, Zhun and Zheng, Liang and Kang, Guoliang and Li, Shaozi and Yang, Yi},
  booktitle={Proceedings of the AAAI conference on artificial intelligence},
  volume={34},
  number={07},
  pages={13001--13008},
  year={2020}
}

@article{zhang2021delving,
  title={Delving deep into label smoothing},
  author={Zhang, Chang-Bin and Jiang, Peng-Tao and Hou, Qibin and Wei, Yunchao and Han, Qi and Li, Zhen and Cheng, Ming-Ming},
  journal={IEEE Transactions on Image Processing},
  volume={30},
  pages={5984--5996},
  year={2021},
  publisher={IEEE}
}

@inproceedings{kim2025exploiting,
  title={Exploiting Diffusion Prior for Task-driven Image Restoration},
  author={Kim, Jaeha and Oh, Junghun and Lee, Kyoung Mu},
  booktitle={Proceedings of the IEEE/CVF International Conference on Computer Vision},
  pages={10151--10161},
  year={2025}
}

@article{chen2024deep,
  title={Deep compression autoencoder for efficient high-resolution diffusion models},
  author={Chen, Junyu and Cai, Han and Chen, Junsong and Xie, Enze and Yang, Shang and Tang, Haotian and Li, Muyang and Lu, Yao and Han, Song},
  journal={arXiv preprint arXiv:2410.10733},
  year={2024}
}
}

\end{document}